\documentclass[lettersize,journal]{IEEEtran}
\usepackage{amsmath,amsfonts}
\usepackage{algorithmic}
\usepackage{algorithm}
\usepackage{array}
\usepackage[caption=false,font=normalsize,labelfont=sf,textfont=sf]{subfig}
\usepackage{textcomp}
\usepackage{stfloats}
\usepackage{url}
\usepackage{verbatim}
\usepackage{graphicx}
\usepackage{cite}
\usepackage[numbers]{natbib}
\usepackage{orcidlink}
\usepackage{hyperref}     
\usepackage{booktabs}     
\usepackage{nicefrac}     
\usepackage{microtype}      
\usepackage{xcolor} 
\usepackage[table]{xcolor}
\usepackage{wrapfig}
\usepackage{makecell}
\usepackage{tabularx}
\usepackage{comment}
\usepackage{wrapfig}
\usepackage{subcaption}
\usepackage{graphicx}
\usepackage{multirow}
\usepackage{diagbox}
\usepackage{bm}

\begin{document}

\title{Attention-Free and Lightweight Token Reduction for Efficient Vision-Language Models}


\author{Xuanyi Hao\orcidlink{0009-0008-9742-9116},
        Zuoyuan Zhang\orcidlink{0000-0002-8853-0078},
        Zhibo Wang\orcidlink{0000-0002-5804-3279},~\IEEEmembership{Senior Member,~IEEE,} 
        Xiaoyi Pang\orcidlink{0000-0002-2763-2695},    
        Jiahui Hu\orcidlink{0000-0001-8771-7474},
        Jiacheng Du\orcidlink{0009-0006-8703-1388},
        Shuguo Zhuo\orcidlink{0000-0001-5200-7130}
        
\thanks{Xuanyi Hao, Zuoyuan Zhang, Zhibo Wang, Jiacheng Du, Shuguo Zhuo, are with the State Key Laboratory of Blockchain and Data Security, Zhejiang University, Hangzhou 310027, China. (E-mail: {xyhao, zuoyuan2024, zhibowang, jcdu, shuguozhuo}@zju.edu.cn).}  

\thanks{Xiaoyi Pang is with the Hong Kong University of Science and Technology, Hong Kong SAR, China. (E-mail: xypang@whu.edu.cn).}  

\thanks{Jiahui Hu is with the School of Artificial Intelligence, Nanchang Univer
sity, Nanchang 330038, China. (E-mail: jiahuihu@ncu.edu.cn).}  
               
}



\maketitle

\begin{abstract}
Vision-Language Models (VLMs) have achieved strong performance in multimodal understanding, yet remain challenging to deploy on resource-constrained edge devices due to the substantial computational overhead of processing numerous visual tokens.
Token reduction is a promising direction for accelerating VLMs inference, but existing approaches either rely on attention maps that are incompatible with modern acceleration frameworks or depend on computationally intensive pairwise similarity comparisons, which undermine scalability and negate their practical benefits in deployment.
In this paper, we propose an attention-free and lightweight token reduction framework as a plug-and-play module for VLMs, which preserves both important and diverse tokens to produce a compact visual representation.
First, to enable attention-free importance estimation, we adopt an information-theoretic perspective and quantify token information using a novel entropy-based criterion, retaining those with more expressive and less degenerate feature representations.
Second, to ensure diverse visual coverage in a lightweight manner, we introduce a transformation-induced consistency signal where similar tokens yield similar signals, such that sorting by this signal places similar tokens close to each other and enables stride-based selection to produce a diverse token set.
Extensive experiments across multiple VLMs benchmarks demonstrate that our framework achieves a favorable accuracy-efficiency trade-off, maintaining competitive performance under aggressive compression.
\end{abstract}


\section{Introduction} \label{sec:Introduction}

Vision-Language Models (VLMs) have emerged as a powerful paradigm for multimodal understanding, demonstrating remarkable performance on tasks such as visual question answering, image captioning, and multimodal dialogue \citep{gemini, internvl, qwen2_vl, blip}.
By coupling a vision encoder with a large language model (LLM), VLMs are able to align visual representations with linguistic semantics and perform complex cross-modal inference, highlighting their potential for edge and mobile applications.
However, deploying VLMs on resource-constrained edge devices \cite{edge4, edge3, edge2, edge1} remains highly challenging due to their substantial computational overhead.  
A primary contributor to this overhead is the excessive number of visual tokens generated by vision encoders \citep{ViT, clip, siglip, TAP-ViT}, often reaching hundreds or even thousands per image. 
This substantially increases the input sequence length and the quadratic cost of self-attention during LLM inference, imposing significant computational burdens on edge hardware.
Therefore, reducing the number of visual tokens while preserving essential visual information is crucial for enabling scalable and efficient VLM deployment.

Token reduction \citep{FastV, VScan, Adaptinfer, Variation-aware} has thus been recognized as a promising direction to alleviate this bottleneck and accelerate VLM inference by retaining informative tokens while removing redundant ones.
Most existing approaches estimate token importance using attention maps, dropping or merging visual tokens according to their attention weights \citep{SparseVLMs, PyramidDrop, MustDrop, lightvlm, VisionZip, VisPruner}.
Although effective, these methods fundamentally rely on explicit attention computation, rendering them incompatible with modern acceleration frameworks such as FlashAttention \citep{flash_attn, flash_attn_2, flash_attn_3, vllm}, which avoid materializing attention maps.
This incompatibility significantly constrains further inference acceleration in large-scale VLMs.
Other works attempt to avoid this limitation by selecting tokens based on feature similarity \citep{similarity, dymu, holitom, dycoke}.
However, these methods typically require extensive pairwise token comparisons, introducing substantial computational overhead that weakens the practical efficiency gains of token reduction.
These limitations highlight the need for a vision token reduction approach that simultaneously achieves strong effectiveness, high efficiency, and broad compatibility.

In this paper, we aim to develop a vision token reduction framework that operates between the vision encoder and the LLM, enabling efficient token compression while remaining fully compatible with modern acceleration frameworks. 
However, achieving this goal presents several fundamental challenges.
First, determining which visual tokens to retain is inherently challenging due to the dense and highly redundant nature of visual signals.
Without prior knowledge of downstream cross-modal interactions, it is difficult to identify which tokens are truly critical for preserving accurate multimodal reasoning after token reduction.
Second, there exists a fundamental tension between being attention-free and computationally lightweight. 
Attention mechanisms naturally emphasize visually salient tokens through their learned weighting patterns, making attention maps a convenient signal for token importance estimation. 
Without access to such cues, alternative strategies often rely on extensive token interactions, which inevitably introduce substantial computational overhead. 
Designing an attention-free token reduction method that remains computationally lightweight therefore becomes particularly challenging.

To address these challenges, we propose an attention-free and lightweight vision token reduction method, called ALTR.
Our key idea is to perform one-shot token selection by jointly modeling token importance and diversity, ensuring that the retained tokens are both informative and visually representative.
To achieve efficient and reliable importance assessment without attention maps, we revisit the visual tokens themselves and observe that informative tokens tend to exhibit richer and less degenerate feature distributions.
Guided by this insight, we quantify token importance through an information-theoretic perspective, measuring the intrinsic information content of each token based solely on its feature activations.
This allows us to identify and retain the most informative visual tokens in a single pass, using a computation that is independent of attention mechanisms and scales linearly with the number of tokens.
Beyond preserving highly informative tokens, effective vision-language reasoning also requires maintaining sufficient diversity among the retained visual representations, especially when downstream tasks are unknown.
Rather than explicitly comparing tokens pairwise, which would incur prohibitive computational costs, we introduce a transformation-induced consistency signal, where tokens with similar representations exhibit similar change signals when processed through the same network layer.
Sorting tokens by this signal, a scalar value for each token, naturally places similar tokens close to each other, enabling stride-based sampling to select tokens from different regions of the ordered list and thereby produce a diverse token set with negligible overhead compared to pairwise vector comparisons.
Together, the two criteria are integrated through a ratio hyperparameter that allocates tokens between importance-aware selection and diversity-aware selection, producing a compact yet expressive token set for efficient VLM inference.

Our main contributions can be summarized as follows:
\begin{itemize}
    \item We introduce an \textbf{attention-free and lightweight token reduction framework} for Vision-Language Models, enabling efficient visual token compression while remaining fully compatible with modern acceleration frameworks.
    \item We propose a \textbf{joint importance-aware and diversity-aware token reduction mechanism} that combines entropy-guided importance estimation with variation-guided diversity sampling, enabling informative and diverse token retention without relying on attention maps or expensive pairwise token interactions.
    \item  Extensive experiments show that our framework consistently achieves state-of-the-art performance across various token budgets on LLaVA-v1.5 and generalizes well to LLaVA-Next and Qwen2.5-VL. Furthermore, compared to existing methods, our approach incurs lower computational overhead and is fully compatible with modern acceleration frameworks, making it more advantageous for \textbf{deployment on resource-constrained edge devices}.
\end{itemize}

The remainder of this paper is organized as follows. Section~\ref{sec:Related Works} reviews related works on VLMs and existing token reduction strategies. Section~\ref{sec:Preliminary} introduces the internal architecture of VLMs and formulates the token reduction problem. Section~\ref{sec:Methodology} presents the proposed ALTR token reduction framework in detail. Section~\ref{sec:Experiments} provides comprehensive experimental evaluations of our proposed method. Section~\ref{sec:Conclusion and Discussion} concludes the paper and discusses future research directions.

\section{Related Works} \label{sec:Related Works}

\subsection{Vision-Language Models}
Vision-Language Models (VLMs) have witnessed rapid evolution in recent years, significantly advancing the capability of unified multimodal understanding.
Early research in this area primarily focused on aligning visual and textual representations through dual-encoder architectures, such as CLIP-style models \cite{clip, siglip}, where images and texts are encoded independently and matched in a shared embedding space.
While effective for retrieval and classification tasks, these approaches lack fine-grained cross-modal interaction.
To enable deeper multimodal reasoning, subsequent works introduced cross-attention mechanisms that explicitly fuse visual and textual features.
Representative models, such as ViLBERT \cite{vilbert} and UNITER \cite{uniter}, adopt transformer-based architectures with cross-modal attention layers, allowing bidirectional interaction between image regions and textual tokens.
Despite improved performance, these models are typically trained on moderate-scale datasets and are limited in generalization ability.

More recently, the emergence of large language models has reshaped the design of VLMs.
The dominant paradigm shifts toward a modular architecture that couples a pretrained vision encoder with a powerful LLM.
In this framework, visual inputs are first tokenized by a vision backbone and then projected into the language embedding space, enabling the LLM to perform multimodal reasoning in an autoregressive manner.
This design significantly improves scalability and task generalization.
Building upon this paradigm, recent large-scale VLMs, such as Gemini \citep{gemini}, InternVL \citep{internvl}, and Qwen-VL \citep{qwen2_vl}, further enhance performance through large-scale multimodal pretraining and instruction tuning.
These models demonstrate strong capabilities across diverse tasks, including visual question answering, image captioning, and multimodal dialogue, and exhibit emergent reasoning abilities.

Despite these advances, the tokenized representation of visual inputs introduces substantial computational overhead.
Modern vision encoders, particularly ViT-based architectures \citep{ViT}, decompose an image into a large number of visual tokens, often reaching hundreds or even thousands per image, many of which are redundant or less informative.
When these tokens are fed into the LLM, the sequence length increases significantly, resulting in higher memory consumption and quadratic computational complexity in self-attention.
This challenge becomes even more severe in high-resolution or multi-image scenarios.
Consequently, improving the efficiency of visual token processing has become a critical research direction for scalable VLM deployment.

\subsection{Token Reduction for VLMs}
Token reduction has been widely explored in VLMs as an effective strategy for improving computational efficiency \cite{vflowopt, hidrop, todre, tamp, swiftvlm, fitprune}.
A predominant class of methods relies on attention maps to estimate token importance, discarding tokens with low attention weights.
For instance, SparseVLMs \citep{SparseVLMs} utilizes the self-attention matrices within VLMs to identify vision-related textual tokens as raters, and employs their attention responses to estimate the importance of visual tokens.
PyramidDrop \citep{PyramidDrop} leverages the attention between the last instruction token and visual tokens to estimate visual tokens importance and progressively discards redundant tokens at predefined stages.
MustDrop \citep{MustDrop} estimates visual token importance across encoding, prefilling, and decoding stages, and progressively prunes redundant tokens via a dual-attention filtering mechanism.
While these approaches effectively reduce token counts, their reliance on explicit attention computation makes them incompatible with modern acceleration frameworks such as FlashAttention, which avoid materializing attention maps for improved efficiency.

To address this limitation, several works propose attention-free token reduction strategies that identify redundant tokens based on token similarity metrics.
For instance, SAINT \citep{similarity} identifies redundant tokens by modeling token similarity with a graph-based formulation, where highly similar tokens are pruned.
DyMU \citep{dymu} dynamically merges similar visual tokens to reduce redundancy and introduces virtual token unmerging to reconstruct the full attention dynamics.
While eliminating the need for attention maps, such approaches often introduce additional computation due to extensive token interactions, frequently with quadratic complexity in the number of tokens.

Moreover, some works incorporate token reduction mechanisms directly into the model architecture.
These methods typically introduce additional modules \cite{llava_mini, lvpruning} or redesign vision encoders \cite{matryoshka0, matryoshka1} to progressively compress token sequences during forward propagation.
While effective in improving efficiency, these approaches are tightly coupled with model architecture and training procedures.
As a result, they usually require end-to-end retraining or substantial fine-tuning, which limits their applicability to existing pretrained VLMs and hinders direct reuse in practical scenarios.

Therefore, there remains a need for a plug-and-play token reduction framework that can effectively identify key visual tokens, without relying on attention mechanisms or incurring heavy computation.

\section{Preliminary} \label{sec:Preliminary}

\subsection{Architecture of Vision-Language Models}

A typical Vision-Language Model (VLM) comprises three main components:  
(i) a vision encoder $\mathcal{E}_v$ that extracts visual representations from an input image,  
(ii) a projector $\mathcal{P}$ that maps visual features into the embedding space of the language model, and  
(iii) a large language model (LLM) $\mathcal{E}_\ell$ that performs multimodal reasoning and generation.

Given an input image $\mathbf{I}$, the vision encoder produces a sequence of $N$ visual tokens:$\mathbf{X} = \mathcal{E}_v(\mathbf{I}) = \{ \mathbf{x}_i \}_{i=1}^{N}$, where $\mathbf{x}_i \in \mathbb{R}^{D}$. These tokens are then projected by the projector into the LLM embedding space:$\mathbf{Z} = \mathcal{P}(\mathbf{X})$, where $\mathbf{z}_i \in \mathbb{R}^{d_\ell}$ and $d_\ell$ denotes the LLM's hidden dimension. Finally, the LLM consumes the projected visual tokens $\mathbf{Z}$ together with textual inputs to perform multimodal understanding, reasoning, and response generation.

\subsection{Problem Formulation}

The objective of this work is to identify a compact subset of visual tokens that preserves the essential visual content of the original image, while substantially reducing the computational cost associated with processing all $N$ tokens.
Given the full visual token sequence $\mathbf{X} = \{ \mathbf{x}_i \}_{i=1}^{N}$, we aim to select a reduced subset
\begin{equation}
\mathbf{X}^{*} \subset \mathbf{X}, \qquad |\mathbf{X}^{*}| = K \ll N,
\end{equation}
such that the multimodal reasoning performance achieved using $\mathbf{X}^{*}$ closely approximates that obtained with the complete token set $\mathbf{X}$.

Formally, we define a token selection operator $\mathcal{S}$ that maps the original token sequence to a reduced one, i.e., $\mathbf{X}^{*} = \mathcal{S}(\mathbf{X})$.
The token reduction problem can then be formulated as the following constrained optimization:
\begin{equation}
\begin{aligned}
\min_{\mathcal{S}} \quad & |\mathcal{S}(\mathbf{X})| \\
\text{s.t.} \quad & \mathcal{L}\!\left( \mathcal{F}(\mathcal{S}(\mathbf{X})), \mathcal{F}(\mathbf{X}) \right) \le \delta,
\end{aligned}
\end{equation}
where $\mathcal{F}(\cdot)$ denotes a fixed VLM, $\mathcal{L}(\cdot,\cdot)$ measures the discrepancy between the model outputs with and without token reduction, and $\delta$ specifies a tolerable level of performance degradation.

In practice, directly solving the above optimization is intractable.
Consequently, a practical token selection operator $\mathcal{S}$ should satisfy two key desiderata:
(i) prioritizing tokens that carry high information content and are critical for downstream multimodal reasoning, and
(ii) maintaining sufficient semantic diversity among the selected tokens to mitigate redundancy.
These desiderata provide principled guidance for the development of effective and efficient token reduction methods.

\section{Methodology} \label{sec:Methodology}

\begin{figure*}
  \centering
  \includegraphics[width=\textwidth]{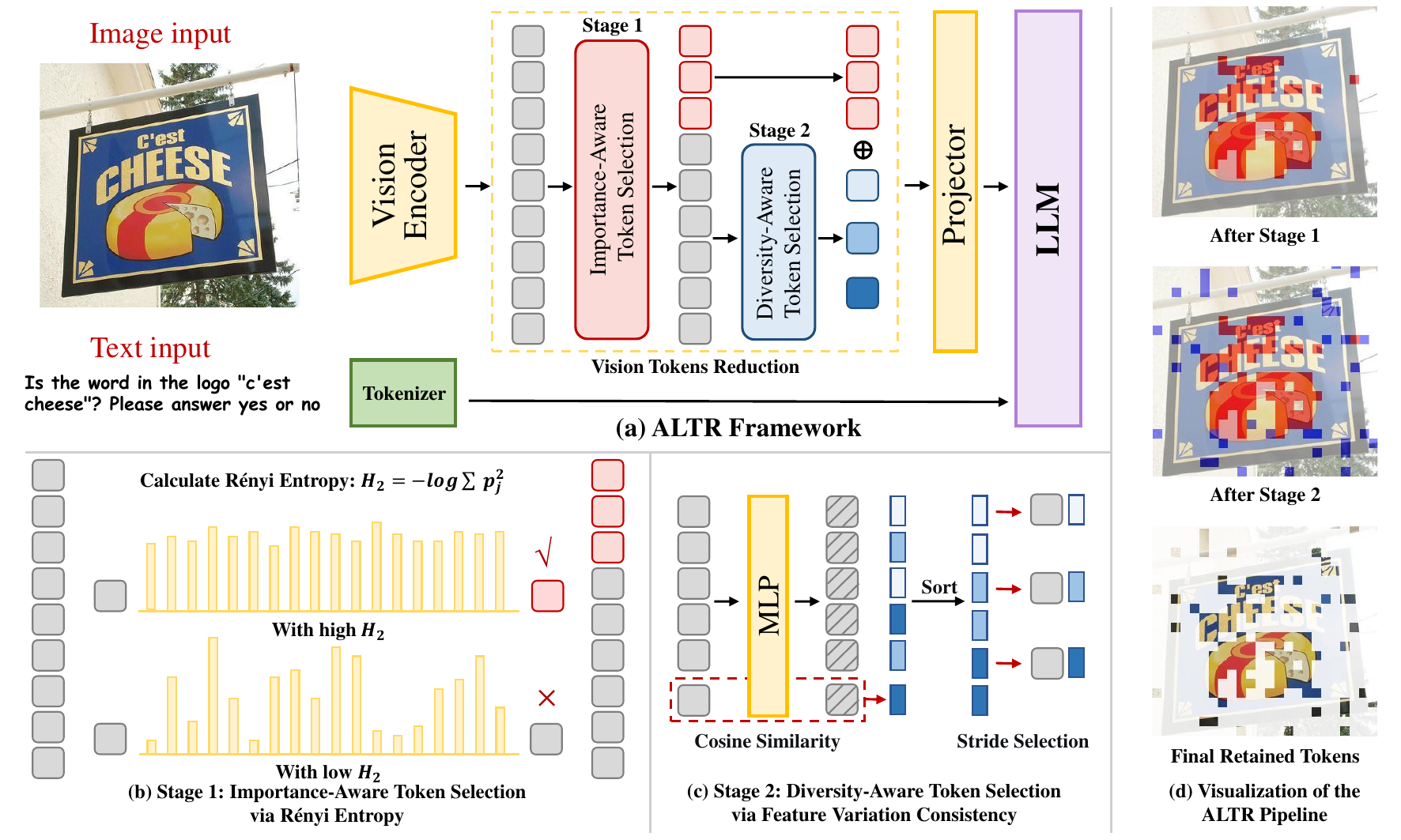}
  \caption{The Framework of ALTR.}
  \label{fig:Framework}
\end{figure*}

We propose an attention-free and lightweight vision token reduction framework ALTR (see Fig.~\ref{fig:Framework}) for Vision-Language Models, which operates between the vision encoder and the LLM and approximates a bi-objective selection problem that balances importance and diversity.
Given an input image, the vision encoder first produces a sequence of visual tokens.
Instead of feeding all tokens into the LLM, we perform a one-shot token reduction to select a compact yet informative subset of visual tokens.
Our method consists of two sequential stages.
In the first stage, we estimate token importance using second-order Rényi entropy, which measures the dispersion of feature activations across dimensions. By interpreting normalized feature magnitudes as a distribution over latent semantic components, tokens with higher entropy are considered to encode richer and less degenerate information.
In the second stage, we leverage the consistency of token-wise transformations induced by the final-layer MLP. The cosine similarity between each token and its transformed representation serves as a lightweight proxy for characterizing transformation sensitivity. Sorting tokens based on this metric organizes them in a latent semantic order, along which stride-based sampling ensures broad coverage of the feature space.
The final selected tokens are then projected into the LLM embedding space and processed jointly with textual tokens for multimodal reasoning.
The two stages are described in detail in the following subsections.

\subsection{Stage I: Importance-Aware Token Selection via Rényi Entropy}

The goal of Stage I is to identify visually informative tokens that contribute substantial semantic content to downstream multimodal reasoning.
Different from approaches that utilize attention maps to guide token selection, we find that tokens with richer and more uniform feature activations tend to encode important visual patterns, whereas tokens with highly concentrated or degenerate activations often correspond to low-information regions.
To formalize this intuition, we interpret each token representation as a composition of latent semantic components distributed across feature dimensions.
Under this view, the magnitude of each feature channel reflects the contribution of a corresponding latent factor.
We therefore normalize the feature values to obtain a discrete distribution over dimensions, enabling a principled measurement of activation dispersion.

Let $\mathbf{X} = \{ \mathbf{x}_i \}_{i=1}^{N}$ denote the visual tokens produced by the vision encoder, where $\mathbf{x}_i \in \mathbb{R}^D$.
For each token $\mathbf{x}_i$, we normalize its feature values along the channel dimension as:
\begin{equation}
p_{i,j} = \frac{|x_{i,j}|}{\sum_{k=1}^{D} |x_{i,k}|},
\end{equation}
where $p_{i,j}$ can be interpreted as a discrete probability distribution over latent semantic components.
The use of absolute values avoids sign cancellation and ensures that activation magnitudes faithfully reflect feature contributions.

We then quantify the concentration of this distribution using the second-order Rényi entropy:
\begin{equation}
H_2(\mathbf{x}_i) = -\log \sum_{j=1}^{D} p_{i,j}^2.
\end{equation}
Compared to Shannon entropy, the second-order Rényi entropy is computationally efficient and places greater emphasis on dominant components, making it particularly suitable for measuring activation concentration in high-dimensional feature spaces.
A higher entropy value corresponds to a more uniform distribution, indicating that the token activates a broader set of latent components and thus encodes richer and less degenerate semantic information.

Based on this criterion, we select the top-$K_1$ tokens with the highest entropy values, forming the importance-aware token set $\mathbf{X}^{imp}$.

\subsection{Stage II: Diversity-Aware Token Selection via Feature Variation Consistency}

While Stage I identifies visually informative tokens, effective vision-language reasoning further requires preserving diverse visual cues, as semantically similar tokens often provide limited complementary information, whereas diversity enables broader contextual coverage and more robust multimodal reasoning.
To this end, we introduce a diversity-aware token selection strategy based on feature variation consistency.

Our core insight is that the MLP-induced representation change serves as an implicit indicator of semantic proximity, as tokens with similar visual content undergo consistent transformation trajectories under the same nonlinear mapping.
This observation allows us to organize tokens in a latent space defined by their transformation responses without explicit pairwise comparisons. 

Formally, for each token $\mathbf{x}_i \in \mathbf{X} \setminus \mathbf{X}^{imp}$, we extract its pre- and post-MLP representations $\mathbf{x}_i'$ and $\mathbf{x}_i''$ from the final-layer MLP (denoted as $f(\cdot)$) of the vision encoder.
The transformation is given by
\begin{equation}
\mathbf{x}_i'' = f(\mathbf{x}_i') = W_2 \, \sigma(W_1 \mathbf{x}_i'),
\end{equation}
where $W_1$ and $W_2$ are weight matrices of the MLP and $\sigma(\cdot)$ is the activation function.
Notably, both $\mathbf{x}_i'$ and $\mathbf{x}_i''$ are obtained from the standard forward pass of the vision encoder, incurring no additional computational overhead.

To quantify the transformation-induced variation, we measure the cosine similarity between the original and transformed representations:
\begin{equation}
s_i = \cos(\mathbf{x}_i', \mathbf{x}_i'') = 
\frac{\mathbf{x}_i'^\top \mathbf{x}_i''}{\|\mathbf{x}_i'\|_2 \|\mathbf{x}_i''\|_2}.
\end{equation}
This scalar metric captures the degree of directional consistency under the MLP transformation. 
More specifically, tokens with similar semantic characteristics tend to yield similar $s_i$ values, whereas semantically distinct tokens are more likely to be dispersed in this transformation-response space. 
Therefore, sorting tokens according to $s_i$ provides an efficient proxy for grouping semantically similar tokens.

Based on this ordering, we perform stride-based sampling to ensure uniform coverage. 
Given a target budget $K_2$, we select tokens at a fixed interval of $(N - |\mathbf{X}^{imp}|)/K_2$ along the sorted sequence. 
This method can be interpreted as uniform sampling in a latent semantic axis induced by transformation responses, effectively preventing the concentration of similar tokens in the final set.

This strategy effectively enhances diversity among the retained tokens, yielding a diversity-aware token set $\mathbf{X}^{div}$.

\subsection{Overall Token Reduction Pipeline}

The overall pipeline formalizes the token reduction process as a structured selection problem, as illustrated in Algorithm~\ref{alg:ALTR}. Given a set of $N$ visual tokens, our method constructs the reduced token set $\mathbf{X}^*$ by combining two complementary subsets: an importance-aware subset $\mathbf{X}^{imp}$ and a diversity-aware subset $\mathbf{X}^{div}$.
The importance-aware subset is obtained by ranking tokens according to their Rényi entropy and selecting the top-$K_1$ tokens. The diversity-aware subset is drawn from the remaining tokens via transformation consistency-based ordering followed by stride-based sampling, yielding $K_2$ additional tokens.
The final reduced token collection is given by:
\begin{equation}
\mathbf{X}^{*} = \mathbf{X}^{imp} \cup \mathbf{X}^{div}, \qquad |\mathbf{X}^{*}| = K_1 + K_2 = K,
\end{equation}
where the balance between the two subsets is controlled by $\lambda$, with $K_1 = \lambda K$ and $K_2 = (1-\lambda)K$. 

Notably, the entire pipeline is training-free and does not rely on attention maps or pairwise token interactions.  
As a result, our method is computationally efficient, scalable, and readily compatible with modern acceleration frameworks.

\begin{algorithm}[t]
\caption{VLMs Token Reduction Pipeline}
\label{alg:ALTR}
\begin{algorithmic}[1]
\REQUIRE Visual tokens $\mathbf{X} = \{\mathbf{x}_i\}_{i=1}^{N}$, target budget $K$, balance factor $\lambda \in [0,1]$
\ENSURE Reduced token set $\mathbf{X}^*$ with $|\mathbf{X}^*| = K$

\STATE $K_1 \gets \lfloor \lambda K \rfloor$, $K_2 \gets K - K_1$
\STATE $\mathbf{X}^{imp} \gets \emptyset$, $\mathbf{X}^{div} \gets \emptyset$ 

\vspace{0.5em}
\COMMENT{Stage I: Importance-Aware Token Selection}
\FOR{each token $\mathbf{x}_i \in \mathbf{X}$}
    \STATE Normalize feature magnitudes: 
    \[
    p_{i,j} = \frac{|x_{i,j}|}{\sum_{k=1}^{D} |x_{i,k}|}
    \]
    \STATE Compute the second-order Rényi entropy:
    \[
    H_2(\mathbf{x}_i) = -\log \sum_{j=1}^{D} p_{i,j}^2
    \]
\ENDFOR
\STATE $\mathbf{X}^{imp} \gets \text{Top-}K_1$ tokens by $H_2(\cdot)$

\vspace{0.5em}
\COMMENT{Stage II: Diversity-Aware Token Selection}
\STATE $\mathbf{X}^{rem} \gets \mathbf{X} \setminus \mathbf{X}^{imp}$
\FOR{each $\mathbf{x}_i \in \mathbf{X}^{rem}$}
    \STATE Obtain pre-MLP representation $\mathbf{x}_i'$ and post-MLP representation $\mathbf{x}_i''$ from the final layer of vision encoder
    \STATE Compute similarity score:
    \[
    s_i = \cos(\mathbf{x}_i', \mathbf{x}_i'') = \frac{\mathbf{x}_i'^\top \mathbf{x}_i''}{\|\mathbf{x}_i'\|_2 \|\mathbf{x}_i''\|_2}.
    \]
\ENDFOR
\STATE Sort $\mathbf{X}^{rem}$ by $s_i$ in ascending order
\STATE Compute stride $t \gets \max\bigl(1, \lfloor |\mathbf{X}^{rem}| / K_2 \rfloor\bigr)$
\STATE Select $\mathbf{X}^{div}$ as $\{ \mathbf{x}_{i} \mid i = 1, 1+t, \dots \}$ up to $K_2$ tokens

\vspace{0.5em}
\STATE $\mathbf{X}^* \gets \mathbf{X}^{imp} \cup \mathbf{X}^{div}$
\STATE \textbf{return} $\mathbf{X}^*$
\end{algorithmic}
\end{algorithm}

\section{Experiments} \label{sec:Experiments}

In this section, we conduct comprehensive experiments to evaluate the proposed method from multiple perspectives, including effectiveness, efficiency, compatibility, and practical deployability. 
We begin by introducing the experimental setup, covering the models, datasets, baselines, and implementation details. 
We then present a thorough performance analysis to assess the effectiveness of our method in terms of task accuracy. 
Subsequently, we conduct an efficiency analysis to evaluate the computational cost and latency reduction achieved by our approach. 
Next, we investigate the compatibility of our method with modern system-level acceleration techniques, demonstrating its potential for integration into practical inference pipelines. 
We further validate the real-world applicability of our approach through testbed evaluation on a deployed inference system, highlighting its effectiveness under practical deployment scenarios. 
Finally, we conduct ablation and hyperparameter analyses to evaluate the contribution of key components and justify our design choices.

\begin{table*}[th!]
\centering
\fontsize{8.5pt}{11pt}\selectfont
\caption{\textbf{Performance on LLaVA-v1.5.} Values outside parentheses denote the original accuracy (\%) on TextVQA, ScienceQA, and GQA, or the original total score on MME. Values in parentheses denote the corresponding performance relative to the vanilla model, whose performance is normalized to 100\%. The average is computed over the normalized results.}
\label{tab:v1_5_res}

\newcolumntype{C}{>{\centering\arraybackslash}X}
\begin{tabularx}{\textwidth}{c|CCCC|C}
\toprule \diagbox[width=9em]{Method}{Dataset} & MME & TextVQA & ScienceQA & GQA & Avg. \\ \midrule

\rowcolor[RGB]{225,240,255}
\multicolumn{6}{c}{\textit{Upper Bound, 576 tokens }(100\%)} \\ 
LLaVA-v1.5-7B  & 1861.59 (100\%) & 58.21\% (100\%) & 70.17\% (100\%) & 59.27\% (100\%) & 100\% \\
\midrule

\rowcolor[RGB]{225,240,255}
\multicolumn{6}{c}{\textit{Retain 192 Tokens }($\downarrow$66.7\%)} \\ 
SparseVLMs (ICML 25) & 1776.42 (95.42\%) & 57.75\% (99.21\%) & 69.82\% (99.50\%) & 59.44\% (100.29\%) & 98.61\% \\
PyramidDrop (CVPR 25)  & 1770.23 (95.09\%) & 56.20\% (96.55\%) & 70.03\% (99.80\%) & 57.16\% (96.44\%) & 96.97\% \\
MustDrop (arXiv)  & 1751.11 (94.06\%) & 56.25\% (96.63\%) & 69.98\% (99.50\%) & 57.03\% (96.22\%) & 96.66\% \\
\midrule
VisionZip (CVPR 25)  & 1750.38 (94.03\%) & 57.45\% (98.69\%) & 69.68\% (99.30\%) & 59.28\% (100.02\%) & 98.01\% \\
VisPruner (ICCV 25)  & 1,765.68 (94.85\%) & 57.62\% (98.99\%) & 69.77\% (99.43\%) & 59.33\% (100.10\%) & 98.34\% \\
\rowcolor{gray!25}
\textbf{ALTR (Ours)}& 1,808.74 (97.16\%) & 57.64\% (99.02\%) & 69.65\% (99.26\%) & 59.27\% (100.00\%) & \textbf{98.86\%} \\
\midrule

\rowcolor[RGB]{225,240,255}
\multicolumn{6}{c}{\textit{Retain 128 Tokens }($\downarrow$77.8\%)} \\
SparseVLMs (ICML 25)& 1768.26 (94.99\%) & 56.70\% (97.41\%) & 69.77\% (99.43\%) & 58.32\% (98.40\%) & 97.55\% \\
PyramidDrop (CVPR 25) & 1,645.33  (88.38\%) & 55.11\%  (94.67\%) & 70.01\%  (99.77\%) & 55.22\%  (93.17\%) & 94.00\% \\
MustDrop (arXiv)& 1681.17 (90.31\%) & 56.10\% (96.38\%) & 69.87\% (99.57\%) & 56.17\% (94.77\%) & 95.26\% \\
\midrule
VisionZip (CVPR 25) & 1,749.71 (93.99\%) & 56.64\% (97.30\%) & 69.82\% (99.50\%) & 58.03\% (97.91\%) & 97.18\% \\
VisPruner (ICCV 25) & 1,764.88 (94.80\%) & 57.15\% (98.18\%) & 69.75\% (99.40\%) & 58.43\% (98.58\%) & 97.74\% \\
\rowcolor{gray!25}
\textbf{ALTR (Ours)} & 1,791.51 (96.24\%) & 56.81\% (97.59\%) & 69.87\% (99.57\%) & 58.32\% (98.40\%) & \textbf{97.95\%} \\
\midrule

\rowcolor[RGB]{225,240,255}
\multicolumn{6}{c}{\textit{Retain 64 Tokens }($\downarrow$88.9\%)} \\
SparseVLMs (ICML 25) & 1,591.23 (85.48\%) & 53.47\% (91.86\%) & 70.29\% (100.17\%) & 53.74\% (90.67\%) & 92.04\% \\
PyramidDrop (CVPR 25) & 1,250.47 (67.17\%) & 46.92\% (80.60\%) & 69.56\% (99.13\%) & 43.79\% (73.88\%) & 80.20\% \\
MustDrop (arXiv) & 1,535.76 (82.50\%) & 54.50\% (93.63\%) & 69.91\% (99.63\%) & 52.71\% (88.93\%) & 91.17\% \\
\midrule
VisionZip (CVPR 25) & 1,667.02 (89.55\%) & 54.90\% (94.31\%) & 69.96\% (99.70\%) & 54.97\% (92.75\%) & 94.08\% \\
VisPruner (ICCV 25) & 1,656.33 (88.97\%) & 55.66\% (95.62\%) & 69.79\% (99.46\%) & 55.31\% (93.32\%) & 94.34\% \\
\rowcolor{gray!25}
\textbf{ALTR (Ours)} & 1,699.18 (91.28\%) & 54.60\% (93.80\%) & 69.84\% (99.53\%) & 55.40\% (93.47\%) & \textbf{94.52\%} \\

\bottomrule
\end{tabularx}
\end{table*}

\subsection{Experimental Setup}

\textbf{Models and Datasets.} Following prior work, we implement our method on the LLaVA framework\cite{llava, llava_next} and evaluate it on four widely used benchmarks: MME~\cite{mme}, TextVQA~\cite{textvqa}, ScienceQA~\cite{sqa}, and GQA~\cite{gqa}. These benchmarks cover diverse capabilities including perception, OCR, scientific reasoning, and compositional visual reasoning. We report accuracy (\%) on TextVQA, ScienceQA, and GQA, and the total score on MME, following the official evaluation protocols. For more intuitive assessment, we present the results by percentage format for comparative analysis, and the accuracy or score of the vanilla model with the 100\% upper limit. In addition, we conduct preliminary experiments on the Qwen2.5-VL~\cite{qwen25_vl} series to assess the generalizability of our method across different VLM architectures.

\textbf{Baseline.} We compare our method with five representative and influential baselines, which can be broadly categorized into \textit{Inner-LM} and \textit{Before-LM} approaches. Inner-LM methods, including SparseVLMs\cite{SparseVLMs}, PyramidDrop\cite{PyramidDrop}, and MustDrop\cite{MustDrop}, reduce visual tokens within the language model during inference, typically guided by attention maps. In contrast, Before-LM methods, such as VisionZip\cite{VisionZip} and VisPruner\cite{VisPruner}, prune or compress visual features prior to feeding them into the language model. 
These baselines represent commonly adopted strategies for visual token reduction and provide strong benchmarks for evaluating the effectiveness of related works.

\subsection{Performance Analysis}

\textbf{Results on LLaVA-v1.5.}
We evaluate the proposed ALTR method on LLaVA-v1.5~\cite{llava}, a widely adopted setting for visual token pruning. Experiments are conducted across multiple benchmarks, and all results are reported in percentage format with the full-token setting as the 100\% reference baseline. Following prior work~\cite{FastV, VScan, SparseVLMs, PyramidDrop}, we consider three retained vision-token budgets: 192, 128, and 64, corresponding to 66.7\%, 77.8\%, and 88.9\% token reduction, respectively.

When reducing the number of visual tokens from 576 to 192, ALTR preserves 98.86\% of the original average performance without any additional training, incurring only a 1.14\% drop compared with the full-token baseline. Under this moderate pruning setting, ALTR achieves the best overall performance among all compared methods. Specifically, ALTR achieves average performance retention gains of 0.85\%, 0.52\%, 0.25\%, 1.89\%, and 2.20\% over VisionZip, VisPruner, SparseVLMs, PyramidDrop, and MustDrop, respectively, with the largest improvements observed against the latter two. These results indicate that ALTR can remove two-thirds of the visual tokens while almost fully preserving the original multimodal reasoning capability.

For the 128-token setting, where only 22.2\% of the original visual tokens are retained, ALTR still maintains strong performance robustness. It achieves an average performance retention of 97.95\%, corresponding to only a 2.05\% degradation from the full-token baseline. Compared with existing competitive Before-LM approaches, ALTR improves over VisionZip by 0.77\% and VisPruner by 0.21\% in average retention. In contrast, several previous  Inner-LM methods suffer more visible performance drops under this stronger compression ratio, with SparseVLMs, PyramidDrop and MustDrop retaining only 97.55\%, 94.00\% and 95.26\% of the baseline performance, respectively. This demonstrates that ALTR remains effective even when the visual-token budget is further reduced from 192 to 128, showing better stability across benchmarks under aggressive pruning.

Under the most challenging 64-token setting, only 11.1\% of the visual tokens are preserved, meaning that the full-token baseline uses nine times more visual tokens. In this extreme reduction regime, conventional token reduction methods exhibit substantial accuracy degradation. For example, SparseVLMs, PyramidDrop, and MustDrop retain 92.04\%, 80.20\%, and 91.17\% of the original average performance, respectively. By contrast, ALTR preserves 94.52\% of the full-token performance, corresponding to only a 5.48\% average reduction despite the severe compression. ALTR also surpasses VisionZip and VisPruner by 0.44\% and 0.18\%. These results highlight the strong robustness of ALTR under extreme token sparsification and suggest that it can effectively identify and retain the most informative visual tokens even with a highly constrained token budget.

These results verify the effectiveness of ALTR's core design. ALTR does not rely on attention maps and extensive token interactions, but instead selects tokens from the intrinsic statistics of the vision encoder. The second-order R\'enyi entropy helps identify tokens with richer and less degenerate feature activations, while the final-layer MLP transformation provides a lightweight signal to select more diverse tokens. The consistent gains across 192, 128, and 64 retained tokens show that ALTR's advantage comes from exploiting the vision model's own feature distribution and transformation behavior, making it both effective and plug-and-play for VLM token reduction.

\begin{figure}[htbp]
    \centering
    \includegraphics[width=0.48\textwidth]{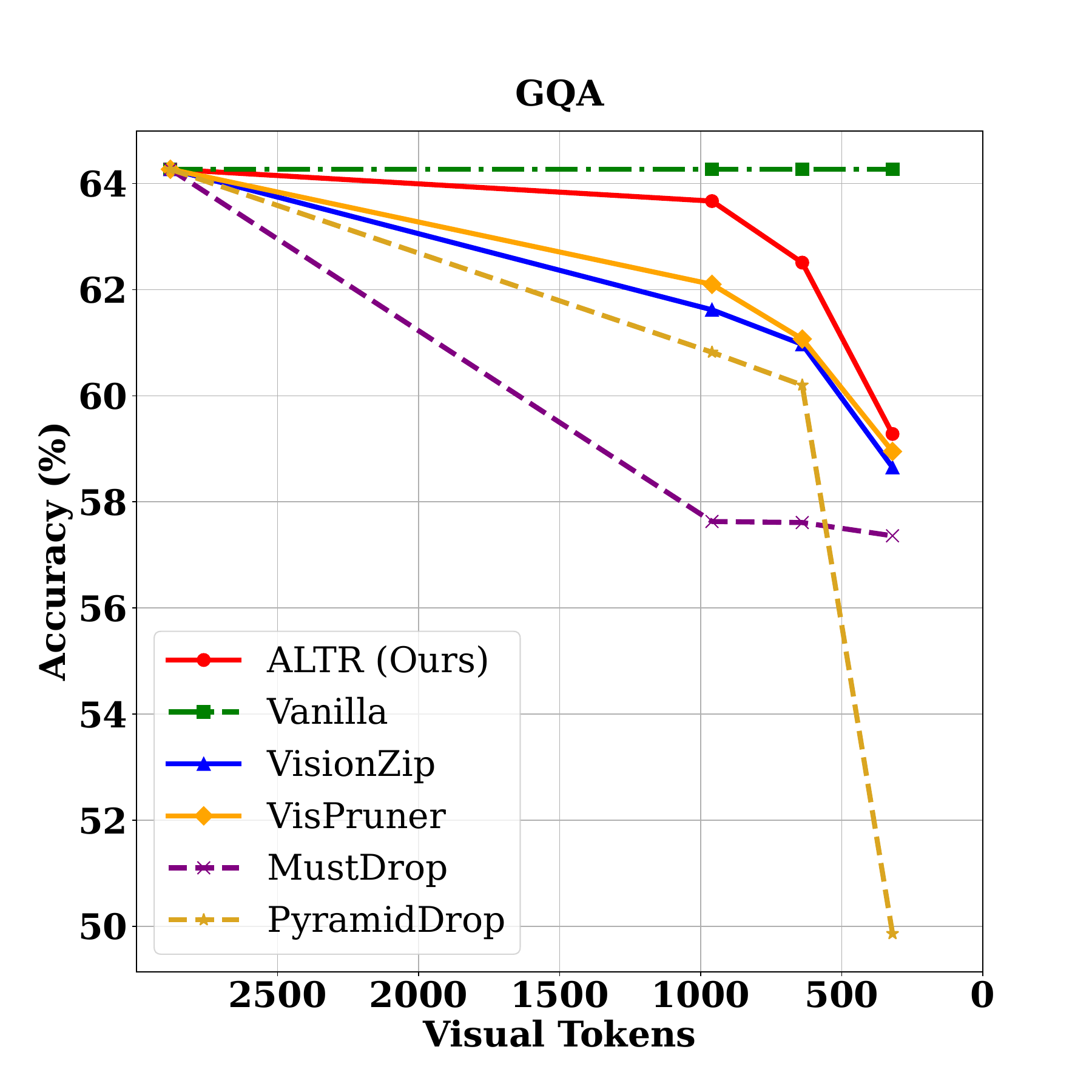}
    \caption{Performance on LLaVA-Next}
    \label{fig:llava_next}
\end{figure}

\textbf{Results on other VLM.}
We further evaluate ALTR on the high-resolution VLM LLaVA-NeXT~\cite{llava_next} on the GQA dataset. LLaVA-NeXT partitions each image into five sub-images, resulting in a substantially larger number of visual tokens and posing a more challenging  visual token reduction problem. As illustrated in Fig.~\ref{fig:llava_next}, the original Vanilla model achieves an accuracy of 64.27\% without token reduction, which serves as an upper-bound performance reference. Across three token retention configurations—33.3\% (960 tokens), 22.2\% (640 tokens), and 11.1\% (320 tokens)—ALTR consistently achieves the best accuracy among all compared methods, demonstrating strong scalability and robustness under aggressive token compression.

Specifically, with 960 tokens, ALTR attains 63.67\% accuracy, which is very close to the Vanilla performance (64.27\%) while significantly outperforming VisionZip (61.62\%), VisPruner (62.10\%), PyramidDrop (60.82\%), and MustDrop (57.63\%), without requiring any additional training. When further reducing the token budget to 640 tokens, ALTR maintains a high accuracy of 62.51\%, preserving most of the original performance and still surpassing all baselines, indicating its effectiveness in retaining critical visual information under moderate compression. Notably, under the most extreme setting of 320 tokens, ALTR achieves 59.28\% accuracy, demonstrating graceful degradation and significantly outperforming PyramidDrop (49.86\%) while consistently exceeding VisionZip (58.65\%), VisPruner (58.95\%), and MustDrop (57.36\%).

We also conduct experiments on Qwen2.5-VL-7B\cite{qwen25_vl} on the TextVQA dataset under the same reduction configurations. Similar to the observations on LLaVA-NeXT, ALTR consistently preserves strong performance across both high and low visual-token settings. Specifically, when retaining 33.3\% visual tokens, corresponding to about 450 tokens, ALTR achieves 82.45\% accuracy, outperforming VisionZip by 0.95\%. When the retained ratio is further reduced to 22.2\%, about 300 tokens, ALTR still obtains 78.94\% accuracy, improving over VisionZip by 1.97\%. Notably, under the most aggressive compression setting—retaining only 11.1\% of visual tokens, roughly 150 tokens—ALTR achieves 72.19\% accuracy, exceeding VisionZip by 5.48\%, demonstrating its effectiveness and adaptability in token-constrained scenarios.

\subsection{Efficiency Analysis}

We evaluate the efficiency of the token reduction module on LLaVA-v1.5-7B using the MME dataset, with all experiments conducted on a single NVIDIA RTX 5880. For a fair comparison, we select VisionZip and VisPruner as representative baselines, since they follow the same \textit{Before-LM} reduction paradigm as ours, where visual tokens are compressed before being fed into the language model. This setting ensures that all compared methods operate at the same stage of the VLM inference pipeline and are evaluated under the same token budget, i.e., 192 retained visual tokens with 64 importance tokens and 128 diversity tokens.

We first analyze the time complexity of different compression modules when sorting and indexing operations are considered. Let \(N\) denote the number of original visual tokens and \(D\) denote the hidden dimension of each visual token. VisionZip is dominated by token similarity computation and contextual merging, resulting in \(\mathcal{O}(N^2D)\) complexity. VisPruner performs iterative token-token similarity computation with a constant deletion step, leading to \(\mathcal{O}(N^3D)\) complexity. In contrast, our method only introduces token-wise scoring over the hidden dimension and sorting, yielding \(\mathcal{O}(ND + N\log N)\) time complexity. This lower complexity comes from the fact that our method avoids both pairwise token-token similarity computation and dense token merging.

In addition to complexity analysis, we measure the additional computational overhead introduced by each \textit{Before-LM} compression module, including both FLOPs and CUDA time. The FLOPs are computed by summing the floating-point arithmetic operations required by each compression module. CUDA time is measured using the PyTorch \cite{pytorch} Profiler, which records the execution time of GPU kernels during the compression process. Non-arithmetic operations such as \texttt{topk}, \texttt{argsort}, \texttt{argmax}, masking, gather/scatter, and index movement are not included in FLOPs, but are reflected in the measured CUDA time.

As shown in Table~\ref{tab:flops_cuda_time}, our method achieves the lowest complexity and introduces only 4.13M additional FLOPs, which is substantially lower than VisionZip (106.91M) and VisPruner (2880.35M). In terms of actual runtime, our method also achieves the lowest CUDA time, requiring only 0.579 ms, compared with 0.884 ms for VisionZip and 7.151 ms for VisPruner. Although FLOPs do not capture all non-arithmetic overheads, the CUDA time results confirm that our method remains efficient in practice. These results demonstrate that our design provides a favorable trade-off between reduction effectiveness and computational overhead.

\begin{table}[htbp]
\centering
\caption{Comparison of Time Complexity, FLOPs, and CUDA Time}
\label{tab:flops_cuda_time}

\begin{tabular}{lccc}

\toprule
Method & Time Complexity  & FLOPs (M) & CUDA Time (ms) \\
\midrule
VisionZip & $\mathcal{O}(N^2D)$ & 106.91 & 0.884 \\
VisPruner & $\mathcal{O}(N^3D)$ & 2880.35 & 7.151 \\
\rowcolor{gray!25}
\rule[-0.6ex]{0pt}{2.8ex}
\textbf{Ours} & $\bm{\mathcal{O}(ND+N\log N)}$ & \textbf{4.13} & \textbf{0.579} \\
\bottomrule
\end{tabular}
\end{table}

\subsection{Compatibility Analysis}

To evaluate the compatibility of our method with modern system-level acceleration techniques, we conduct experiments on Qwen2.5-VL under a fixed token budget with two different attention implementations. Specifically, we consider two deployment configurations: (1) the baseline configuration, where both the vision encoder and the LLM adopt the standard eager attention implementation; and (2) the accelerated configuration, where FlashAttention-2 \cite{flash_attn_2} is enabled for both the vision encoder and the LLM. This comparison aims to investigate whether our method can effectively preserve the benefits of optimized attention kernels in practical VLM inference scenarios.

As reported in Table~\ref{tab:flash_compatibility}, our method is fully compatible with FlashAttention-2. Compared with the eager implementation, adopting FlashAttention-2 across both the vision and language components significantly improves the overall inference efficiency. The mean vision encoder latency is reduced from 179.632 ms to 35.567 ms, achieving approximately a 5$\times$ speedup. Meanwhile, the mean prefill latency decreases from 214.242 ms to 71.481 ms, demonstrating that kernel-level acceleration in individual modules can effectively translate into end-to-end performance improvements.
Furthermore, the peak memory consumption is reduced from 16.388 GB to 15.622 GB, indicating that FlashAttention-2 provides additional memory savings by avoiding the materialization of large intermediate attention tensors. These results demonstrate that our attention-free design is compatible with efficient attention kernels and can effectively preserve system-level acceleration benefits in practical VLM inference.

\begin{table}[htbp]
\centering
\caption{Compatibility with FlashAttention-2 on Qwen2.5-VL-7B.}
\label{tab:flash_compatibility}
\begin{tabular}{lcccc}
\toprule
Vision Attn. & Vision Time (ms) & Prefill (ms) & Peak Mem. (GB) \\
\midrule
Eager & 179.632 & 214.242 & 16.388 \\
FlashAttn2 & 35.567 & 71.481 & 15.622 \\
\bottomrule
\end{tabular}
\end{table}

\subsection{Testbed Evaluation}

To further assess the practical deployability of visual-token reduction on edge platforms, we conduct an OOM-boundary study on an NVIDIA Jetson AGX Orin, which serves as our edge testbed. Jetson AGX Orin is representative of modern edge AI platforms: it provides GPU acceleration for on-device inference while operating under substantially tighter memory and resource constraints than datacenter GPUs. This makes it a suitable platform for studying the memory bottlenecks that arise when deploying VLMs in realistic edge scenarios.

Our testbed is configured to emulate practical deployments where the VLM process does not have exclusive access to all system memory. In real edge applications, memory is often shared by the operating system, camera or sensor pipelines, preprocessing modules, user applications, and other background services. To capture this constraint in a controlled manner, we impose fixed available-memory budgets on the VLM process and evaluate whether each method can complete long-response generation without exceeding the budget. All methods are evaluated under identical conditions, including batch size 1, eager attention, and the same image-prompt inputs, so that differences in memory feasibility can be attributed to the visual-token reduction strategy rather than changes in the execution setting.

We set the target generation length using the \texttt{max\_new\_tokens} parameter and gradually increase it with a step size of 256 tokens. We then report the maximum OOM-free response length, defined as the largest tested target length that can be completed successfully without CUDA OOM, system-level OOM, process termination, or incomplete execution.

Table~\ref{tab:jetson_16g_oom} reports the results under a 16GB memory budget. Without visual-token reduction, full-token inference successfully completes all tested lengths up to 768 generated tokens, while the first failure occurs at 1024 tokens during the LLM prefill stage. In contrast, visual-token reduction substantially expands the feasible generation range. When retaining 128 or 64 visual tokens, ALTR successfully completes all tested lengths up to 1536 generated tokens, with the first failure observed at 1792 tokens. VisionZip and VisPruner exhibit the same maximum feasible generation length under the same retained-token budgets, indicating that reducing the visual-token prefix is effective for improving edge memory feasibility.

Importantly, this memory-feasibility gain does not come at the cost of heavy reduction-stage overhead. Echoing the efficiency results in Table~\ref{tab:flops_cuda_time}, ALTR remains lightweight on the Jetson platform, introducing 69.9 ms of reduction latency at the 128-token budget, compared with 98.5 ms for VisionZip and 289.3 ms for VisPruner. This suggests that ALTR can improve edge memory feasibility while maintaining low reduction overhead. Notably, SparseVLMs fails during model loading under the 16GB budget, suggesting that methods with additional model-side components may require a higher minimum memory budget before inference can start. This makes them less suitable for memory-constrained edge devices, where the available memory for a single VLM process is often limited.

\begin{table}[t]
\centering
\caption{OOM-free generation boundary on Jetson AGX Orin under a 16GB available-memory budget. }
\label{tab:jetson_16g_oom}
\footnotesize
\setlength{\tabcolsep}{4pt}
\begin{tabular}{l|cccc}
\toprule
Method & Vision Tokens & Max Length & First Fail & Reduction Time \\
\midrule
Full-token & 576 & 768 & 1024 & -- \\
\midrule
VisionZip & 192 & 1280 & 1536 & 106.3 ms \\
VisPruner & 192 & 1280 & 1536 & 293.2 ms \\
SparseVLMs & 192 & 0 & 256 & -- \\
\rowcolor{gray!25}
\textbf{Ours} & 192 & 1280 & 1536 & 73.3 ms \\
\midrule
VisionZip & 128 & 1536 & 1792 & 98.5 ms \\
VisPruner & 128 & 1536 & 1792 & 289.3 ms \\
SparseVLMs & 128 & 0 & 256 & -- \\
\rowcolor{gray!25}
\textbf{Ours} & 128 & 1536 & 1792 & 69.9 ms \\
\midrule
VisionZip & 64 & 1536 & 1792 & 92.2 ms \\
VisPruner & 64 & 1536 & 1792 & 308.0 ms \\
SparseVLMs & 64 & 0 & 256 & -- \\
\rowcolor{gray!25}
\textbf{Ours} & 64 & 1536 & 1792 & 67.1 ms \\
\bottomrule
\end{tabular}
\end{table}

\subsection{Ablation Analysis}

\begin{figure}[htbp]
    \centering
    \includegraphics[width=\columnwidth]{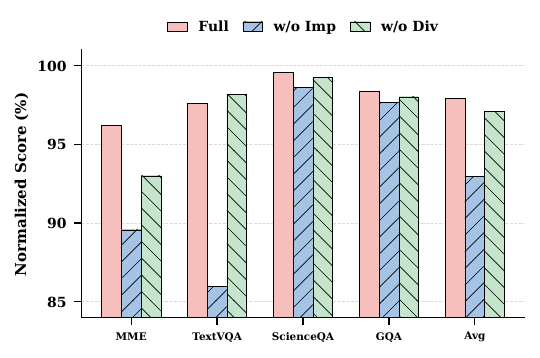}
    \caption{Ablation study results.}
    \label{fig:ablation}
\end{figure}

We conduct ablation experiments on LLaVA-v1.5-7B with 128 retained tokens to investigate the contribution of each proposed component. Specifically, we compare the complete model (\textit{mix}) with two degraded variants, i.e., removing the importance-aware selection module (\textit{w/o Imp}) and removing the diversity-aware selection module (\textit{w/o Div}). The experimental results are presented in Fig.~\ref{fig:ablation}. The complete model achieves the best overall performance, obtaining an average score of 97.95\% across all evaluated benchmarks, which demonstrates the effectiveness of jointly incorporating importance-aware and diversity-aware token selection strategies.

The removal of the importance-aware selection module causes a significant performance degradation, reducing the average score from 97.95\% to 92.96\%. Specifically, the MME score decreases from 96.24\% to 89.57\%, while the TextVQA score drops from 97.59\% to 85.96\%. Meanwhile, consistent declines are observed on ScienceQA and GQA, with the scores decreasing from 99.57\% to 98.63\% and from 98.40\% to 97.67\%, respectively. These results indicate that importance-aware selection is critical for identifying and preserving task-relevant visual tokens. Without this component, the token compression process may discard informative regions and lead to the loss of essential visual semantics, thereby limiting the model's multimodal understanding capability under a restricted token budget.

We further evaluate the effect of the diversity-aware selection module by removing it from the full model. The resulting variant achieves an average score of 97.12\%, which is 0.83\% lower than the complete model. Although minor performance fluctuations can be observed on individual benchmarks, the \textit{w/o Div} variant consistently underperforms the full model in terms of overall performance. In particular, it obtains lower scores on MME (93.00\% vs. 96.24\%), ScienceQA (99.26\% vs. 99.57\%), GQA (98.01\% vs. 98.40\%), and the average score, demonstrating the importance of maintaining sufficient diversity among retained visual tokens. These results suggest that diversity-aware selection provides complementary benefits beyond importance-based filtering by reducing redundant token selection and improving the coverage of visual information. Consequently, the retained token set can preserve richer visual representations and achieve more robust multimodal understanding.

\begin{figure}[t]
\centering

\setlength{\tabcolsep}{2pt} 
\renewcommand{\arraystretch}{1} 

\begin{tabular}{c c c c}

\includegraphics[width=0.24\linewidth]{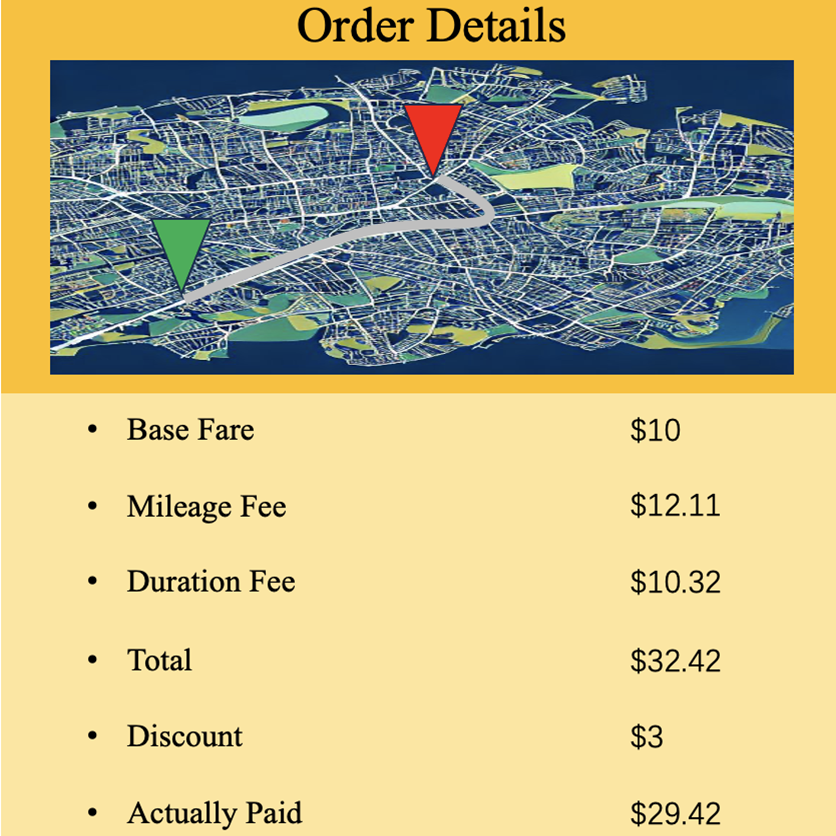} &
\includegraphics[width=0.24\linewidth]{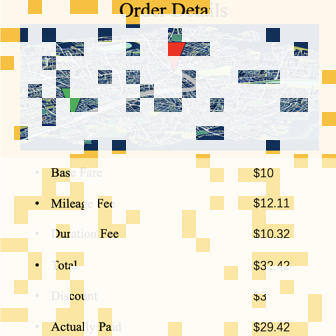} &
\includegraphics[width=0.24\linewidth]{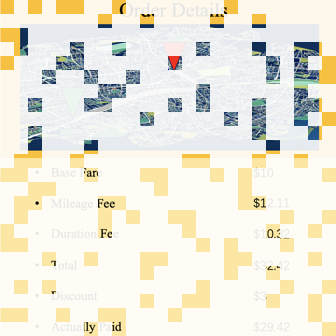} &
\includegraphics[width=0.24\linewidth]{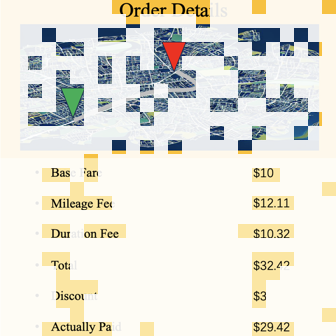} \\

\includegraphics[width=0.24\linewidth]{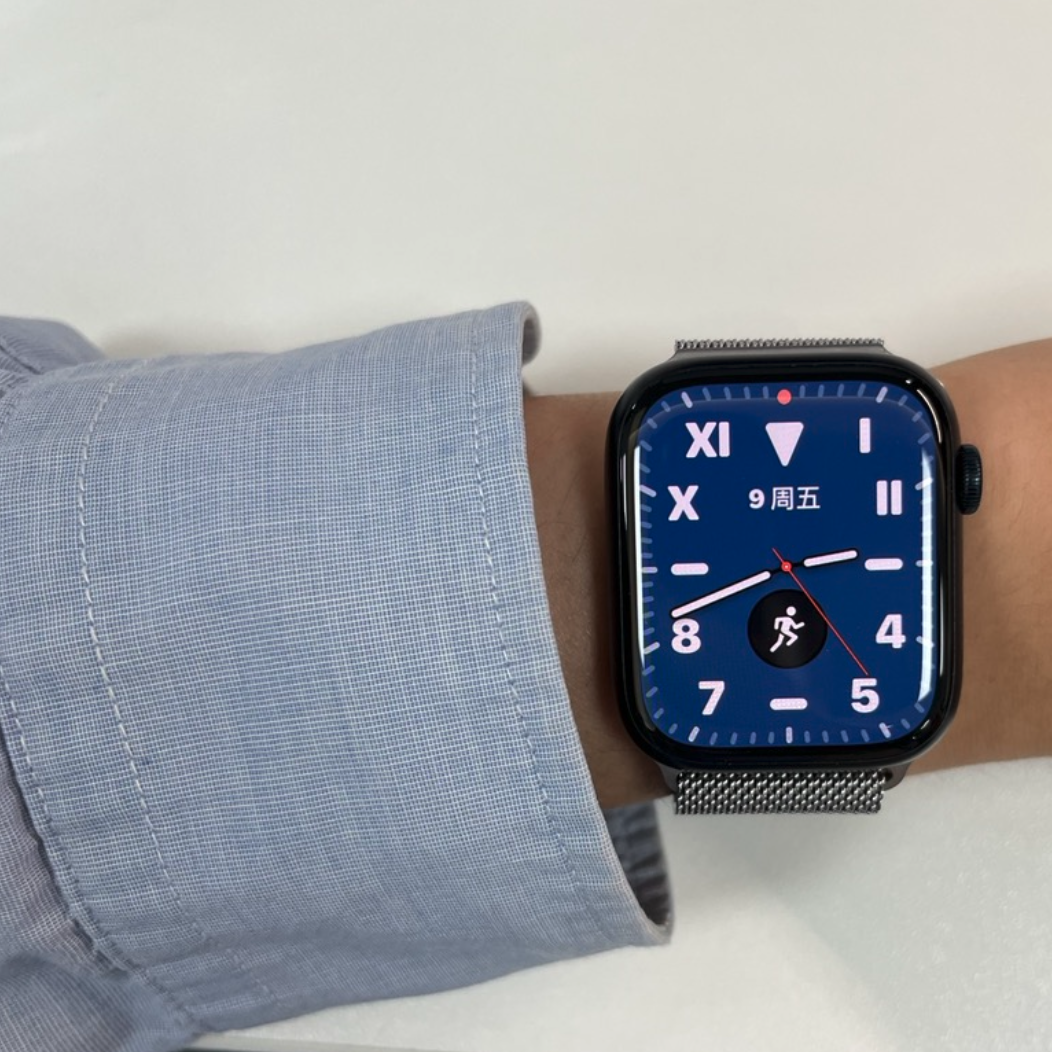} &
\includegraphics[width=0.24\linewidth]{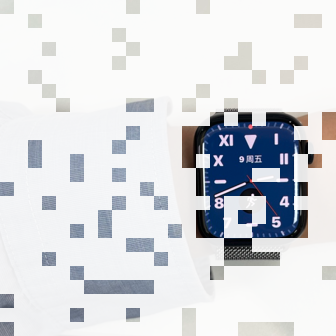} &
\includegraphics[width=0.24\linewidth]{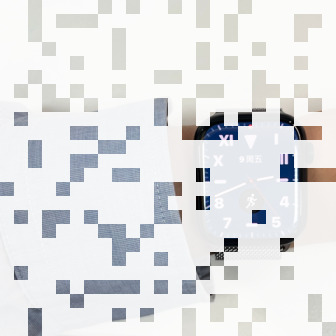} &
\includegraphics[width=0.24\linewidth]{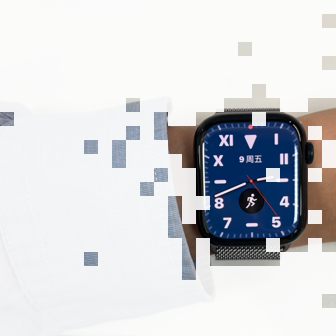} \\

\textbf{Input} & \textbf{Full} & \textbf{w/o Imp} & \textbf{w/o Div} \\

\end{tabular}

\caption{
Qualitative visualization of retained visual tokens in the ablation study. 
}
\label{fig:ablation_vis}
\end{figure}

To further illustrate the effect of each component, we provide qualitative token selection visualizations in Fig.~\ref{fig:ablation_vis}. Compared with the full ALTR, removing the importance-aware stage leads to more scattered and less informative token selection, where several retained tokens fall on background or low-content regions while task-relevant details are weakened. Removing the diversity-aware stage, on the other hand, tends to produce more redundant selections around locally salient regions, resulting in insufficient coverage of the overall visual content. In contrast, the full ALTR selects tokens that better align with informative objects, textual regions, and structural details, while still maintaining a broad spatial and semantic coverage. 

These visual results are consistent with the quantitative ablation results, further showing that the entropy-based importance estimation and variation-based diversity selection play complementary roles in producing a compact yet representative token subset. The importance-aware module mainly ensures the preservation of informative visual content, while the diversity-aware module enhances the completeness and robustness of the retained representation. By combining these two components, the proposed method achieves a favorable balance between token reduction and multimodal reasoning performance.

\subsection{Hyperparameter Analysis}

\begin{figure}[htbp]
    \centering
    \includegraphics[width=\columnwidth]{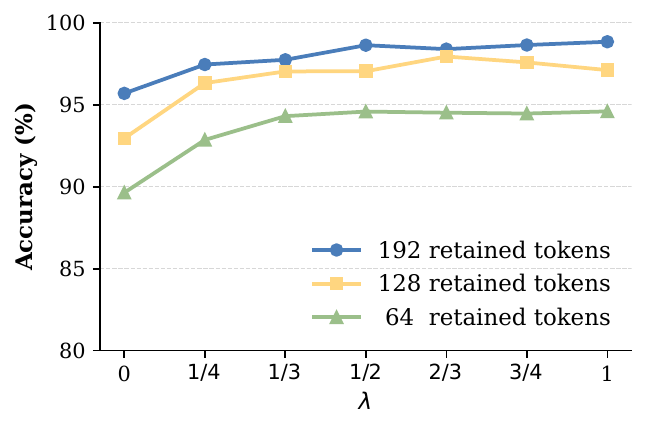}
    \caption{Sensitivity of our method to the balancing coefficient
$\lambda$ under different visual-token budgets.}
    \label{fig:lambda_robustness}
\end{figure}

We analyze the robustness of our method with respect to the balancing hyperparameter $\lambda$, which controls the relative importance of the two components in our token selection strategy. We vary $\lambda$ over a broad range and evaluate the model under different visual token budgets on multiple benchmarks.As shown in Figure~\ref{fig:lambda_robustness}, our method maintains consistently strong performance across different values of $\lambda$. Although the optimal setting may vary slightly across token budgets, the overall performance changes only marginally within a wide range of $\lambda$. This indicates that our method does not rely on careful hyperparameter tuning and remains effective under different configurations. 

In our experiments, to ensure a controlled comparison without method-specific tuning, all approaches involving a balancing hyperparameter adopt the same value of $\lambda$.

\section{Conclusion and Discussion} \label{sec:Conclusion and Discussion}
In this paper, we presented an attention-free and lightweight vision token reduction framework for Vision-Language Models, designed to improve inference efficiency while preserving multimodal reasoning capability.
Our method performs one-shot token selection by jointly modeling token importance and diversity, enabling the construction of a compact yet expressive set of visual tokens for efficient multimodal inference.
The proposed framework avoids reliance on attention maps and expensive pairwise token interactions, making it fully compatible with modern acceleration frameworks and scalable to large visual token sequences.
Extensive experiments demonstrate that our approach consistently achieves substantial token reduction and inference acceleration while maintaining, and in some cases even improving, performance across a wide range of VLM benchmarks.

Beyond its empirical effectiveness, our work offers several conceptual insights.
First, we show that intrinsic properties of visual token representations can provide reliable signals for importance estimation, without requiring attention supervision or additional training.
Second, we highlight that diversity among visual tokens can be promoted through lightweight implicit organization, rather than expensive explicit similarity computation.
Together, these observations suggest a broader perspective on token reduction: efficient selection can be achieved by exploiting structure already present in pretrained representations.

There are several promising directions for future research.
One avenue is to explore adaptive strategies for determining the token budget or dynamically balancing importance and diversity based on task characteristics.
Another direction is to extend the proposed framework to video-language models or multimodal settings with temporal dynamics, where token redundancy is even more pronounced.
We hope this work encourages further investigation into training-free, acceleration-friendly token reduction methods and contributes to the development of more efficient and practical VLMs on edge devices.

\bibliographystyle{IEEEtran}
\bibliography{08References}

@article{gemini,
  title={Gemini: a family of highly capable multimodal models},
  author={Team, Gemini and Anil, Rohan and Borgeaud, Sebastian and Alayrac, Jean-Baptiste and Yu, Jiahui and Soricut, Radu and Schalkwyk, Johan and Dai, Andrew M and Hauth, Anja and Millican, Katie and others},
  journal={arXiv preprint arXiv:2312.11805},
  year={2023}
}

@inproceedings{internvl,
  title={InternVL: Scaling up vision foundation models and aligning for generic visual-linguistic tasks},
  author={Chen, Zhe and Wu, Jiannan and Wang, Wenhai and Su, Weijie and Chen, Guo and Xing, Sen and Zhong, Muyan and Zhang, Qinglong and Zhu, Xizhou and Lu, Lewei and others},
  booktitle = {Proc. of IEEE/CVF CVPR},
  pages={24185--24198},
  year={2024}
}

@article{qwen2_vl,
  title={{Qwen2-VL}:Enhancing Vision-Language Model's Perception of the World at Any Resolution},
  author={Wang, Peng and Bai, Shuai and Tan, Sinan and Wang, Shijie and Fan, Zhihao and Bai, Jinze and Chen, Keqin and Liu, Xuejing and Wang, Jialin and Ge, Wenbin and Fan, Yang and Dang, Kai and Du, Mengfei and Ren, Xuancheng and Men, Rui and Liu, Dayiheng and Zhou, Chang and Zhou, Jingren and Lin, Junyang},
  journal={arXiv preprint arXiv:2409.12191},
  year={2024}
}

@inproceedings{blip,
  title={{BLIP-2}:Bootstrapping language-image pre-training with frozen image encoders and large language models},
  author={Li, Junnan and Li, Dongxu and Savarese, Silvio and Hoi, Steven},
  booktitle = {Proc. of {ICML}},
  pages={19730--19742},
  year={2023},
}

@inproceedings{ViT,
  title={An image is worth 16x16 words: Transformers for image recognition at scale},
  author={Dosovitskiy, Alexey and Beyer, Lucas and Kolesnikov, Alexander and Weissenborn, Dirk and Zhai, Xiaohua and Unterthiner, Thomas and Dehghani, Mostafa and Minderer, Matthias and Heigold, Georg and Gelly, Sylvain and others},
  booktitle={Proc. of ICLR},
  year={2021}
}

@inproceedings{clip,
  title={Learning transferable visual models from natural language supervision},
  author={Radford, Alec and Kim, Jong Wook and Hallacy, Chris and Ramesh, Aditya and Goh, Gabriel and Agarwal, Sandhini and Sastry, Girish and Askell, Amanda and Mishkin, Pamela and Clark, Jack and others},
  booktitle = {Proc. of ICML},
  pages={8748--8763},
  year={2021},
}

@inproceedings{siglip,
  title={Sigmoid loss for language image pre-training},
  author={Zhai, Xiaohua and Mustafa, Basil and Kolesnikov, Alexander and Beyer, Lucas},
  booktitle = {Proc. of IEEE/CVF ICCV},
  pages={11975--11986},
  year={2023}
}

@article{TAP-ViT,
  title={TAP-ViTs: Task-Adaptive Pruning for On-Device Deployment of Vision Transformers},
  author={Wang, Zhibo and Zhang, Zuoyuan and Pang, Xiaoyi and Zhang, Qile and Hao, Xuanyi and Zhuo, Shuguo and Sun, Peng},
  journal={arXiv preprint arXiv:2601.02437},
  year={2026}
}

@inproceedings{FastV,
  title={An image is worth 1/2 tokens after layer 2: Plug-and-play inference acceleration for large vision-language models},
  author={Chen, Liang and Zhao, Haozhe and Liu, Tianyu and Bai, Shuai and Lin, Junyang and Zhou, Chang and Chang, Baobao},
  booktitle = {Proc. of ECCV},
  pages={19--35},
  year={2024},
}

@article{VScan,
  title={{VScan}: Rethinking Visual Token Reduction for Efficient Large Vision-Language Models},
  author={Zhang, Ce and Ma, Kaixin and Fang, Tianqing and Yu, Wenhao
             and Zhang, Hongming and Zhang, Zhisong and Xie, Yaqi
             and Sycara, Katia and Mi, Haitao and Yu, Dong},
  journal={arXiv preprint arXiv:2505.22654},
  year={2025},
}

@inproceedings{Variation-aware,
  title={Variation-aware vision token dropping for faster large vision-language models},
  author={Chen, Junjie and Liu, Xuyang and Wen, Zichen and Wang, Yiyu and Huang, Siteng and Chen, Honggang},
  booktitle={Proc. of IEEE/CVF CVPR},
  pages={3489--3499},
  year={2026}
}

@article{Adaptinfer,
  title={Adaptinfer: Adaptive token pruning for vision-language model inference with dynamical text guidance},
  author={Zhang, Weichen and Zhu, Zhui and Li, Ningbo and Tao, Shilong and Liu, Kebin and Liu, Yunhao},
  journal={arXiv preprint arXiv:2508.06084},
  year={2025}
}

@inproceedings{SparseVLMs,
  title={Sparsevlm: Visual token sparsification for efficient vision-language model inference},
  author={Zhang, Yuan and Fan, Chun-Kai and Ma, Junpeng and Zheng, Wenzhao and Huang, Tao and Cheng, Kuan and Gudovskiy, Denis and Okuno, Tomoyuki and Nakata, Yohei and Keutzer, Kurt and others},
  booktitle={Proc. of ICML},
  pages={74840--74857},
  year={2025}
}

@inproceedings{PyramidDrop,
  title={Pyramiddrop: Accelerating your large vision-language models via pyramid visual redundancy reduction},
  author={Xing, Long and Huang, Qidong and Dong, Xiaoyi and Lu, Jiajie and Zhang, Pan and Zang, Yuhang and Cao, Yuhang and He, Conghui and Wang, Jiaqi and Wu, Feng and others},
  booktitle={Proc. of IEEE/CVF CVPR},
  pages={14593--14603},
  year={2024}
}

@article{MustDrop,
  title={Multi-Stage Vision Token Dropping: Towards Efficient Multimodal Large Language Model},
  author={Liu, Ting and Shi, Liangtao and Hong, Richang and Hu, Yue and Yin, Quanjun and Zhang, Linfeng},
  journal={arXiv preprint arXiv:2411.10803},
  year={2024}
}

@article{lightvlm,
  title={LightVLM: Acceleraing Large Multimodal Models with Pyramid Token Merging and KV Cache Compression},
  author={Hu, Lianyu and Shang, Fanhua and Feng, Wei and Wan, Liang},
  journal={arXiv preprint arXiv:2509.00419},
  year={2025}
}

@inproceedings{VisionZip,
  title={Visionzip: Longer is better but not necessary in vision language models},
  author={Yang, Senqiao and Chen, Yukang and Tian, Zhuotao and Wang, Chengyao and Li, Jingyao and Yu, Bei and Jia, Jiaya},
  booktitle={Proc. of IEEE/CVF CVPR},
  pages={19792--19802},
  year={2025}
}

@inproceedings{VisPruner,
  title={Beyond text-visual attention: Exploiting visual cues for effective token pruning in vlms},
  author={Zhang, Qizhe and Cheng, Aosong and Lu, Ming and Zhang, Renrui and Zhuo, Zhiyong and Cao, Jiajun and Guo, Shaobo and She, Qi and Zhang, Shanghang},
  booktitle={Proc. of IEEE/CVF ICCV},
  pages={20857--20867},
  year={2025}
}

@inproceedings{flash_attn,
  title={Flashattention: Fast and memory-efficient exact attention with io-awareness},
  author={Dao, Tri and Fu, Dan and Ermon, Stefano and Rudra, Atri and R{\'e}, Christopher},
  booktitle={Proc. of NeurIPS},
  volume={35},
  pages={16344--16359},
  year={2022}
}

@inproceedings{flash_attn_2,
  title={Flashattention-2: Faster attention with better parallelism and work partitioning},
  author={Dao, Tri},
  booktitle={Proc. of ICLR},
  volume={2024},
  pages={35549--35562},
  year={2024}
}

@inproceedings{flash_attn_3,
  title={Flashattention-3: Fast and accurate attention with asynchrony and low-precision},
  author={Shah, Jay and Bikshandi, Ganesh and Zhang, Ying and Thakkar, Vijay and Ramani, Pradeep and Dao, Tri},
  booktitle={Proc. of NeurIPS},
  volume={37},
  pages={68658--68685},
  year={2024}
}

@inproceedings{vllm,
  title={Efficient memory management for large language model serving with pagedattention},
  author={Kwon, Woosuk and Li, Zhuohan and Zhuang, Siyuan and Sheng, Ying and Zheng, Lianmin and Yu, Cody Hao and Gonzalez, Joseph and Zhang, Hao and Stoica, Ion},
  booktitle={Proc. of ACM SOSP},
  pages={611--626},
  year={2023}
}

@article{similarity,
  title={Similarity-aware token pruning: Your vlm but faster},
  author={Jeddi, Ahmadreza and Baghbanzadeh, Negin and Dolatabadi, Elham and Taati, Babak},
  journal={arXiv preprint arXiv:2503.11549},
  year={2025}
}

@article{dymu,
  title={Dymu: Dynamic merging and virtual unmerging for efficient vlms},
  author={Wang, Zhenhailong and Purushwalkam, Senthil and Xiong, Caiming and Savarese, Silvio and Ji, Heng and Xu, Ran},
  journal={arXiv preprint arXiv:2504.17040},
  year={2025}
}

@inproceedings{vilbert,
  title={Vilbert: Pretraining task-agnostic visiolinguistic representations for vision-and-language tasks},
  author={Lu, Jiasen and Batra, Dhruv and Parikh, Devi and Lee, Stefan},
  booktitle={Proc. of NeurIPS},
  volume={32},
  pages = {13--23},
  year={2019}
}

@inproceedings{uniter,
  title={Uniter: Universal image-text representation learning},
  author={Chen, Yen-Chun and Li, Linjie and Yu, Licheng and El Kholy, Ahmed and Ahmed, Faisal and Gan, Zhe and Cheng, Yu and Liu, Jingjing},
  booktitle={Proc. of ECCV},
  pages={104--120},
  year={2020},
}

@inproceedings{llava_mini,
  title={Llava-mini: Efficient image and video large multimodal models with one vision token},
  author={Zhang, Shaolei and Fang, Qingkai and Yang, Yang and Feng, Yang},
  booktitle={Proc. of ICLR},
  pages={53285--53310},
  year={2025}
}

@inproceedings{lvpruning,
  title={Lvpruning: An effective yet simple language-guided vision token pruning approach for multi-modal large language models},
  author={Sun, Yizheng and Xin, Yanze and Li, Hao and Sun, Jingyuan and Lin, Chenghua and Batista-Navarro, Riza Theresa},
  booktitle={Findings of NAACL},
  pages={4299--4308},
  year={2025}
}

@inproceedings{matryoshka0,
  title={Matryoshka multimodal models},
  author={Cai, Mu and Yang, Jianwei and Gao, Jianfeng and Lee, Yong Jae},
  booktitle={Proc. of ICLR},
  pages={46254--46272},
  year={2025},
}

@inproceedings{matryoshka1,
  title={Matryoshka query transformer for large vision-language models},
  author={Hu, Wenbo and Dou, Zi-Yi and Li, Liunian H and Kamath, Amita and Peng, Nanyun and Chang, Kai-Wei},
  booktitle={Proc. of NeurIPS},
  volume={37},
  pages={50168--50188},
  year={2024}
}

@inproceedings{llava,
      title={Improved Baselines with Visual Instruction Tuning}, 
      author={Liu, Haotian and Li, Chunyuan and Li, Yuheng and Lee, Yong Jae},
      booktitle={Proc. of IEEE/CVF CVPR},
      pages={26296--26306},
      year={2024},
}

@misc{llava_next,
    title={LLaVA-NeXT: Improved reasoning, OCR, and world knowledge},
    url={https://llava-vl.github.io/blog/2024-01-30-llava-next/},
    author={Liu, Haotian and Li, Chunyuan and Li, Yuheng and Li, Bo and Zhang, Yuanhan and Shen, Sheng and Lee, Yong Jae},
    month={January},
    year={2024}
}

@inproceedings{mme,
  title={MME: A Comprehensive Evaluation Benchmark for Multimodal Large Language Models},
  author={Fu, Chaoyou and Chen, Peixian and Shen, Yunhang and Qin, Yulei and Zhang, Mengdan and Lin, Xu and Yang, Jinrui and Zheng, Xiawu and Li, Ke and Sun, Xing and others},
  booktitle={Proc. of NeurIPS},
  volume={38},
  year={2026}
}

@inproceedings{textvqa,
  title={Towards vqa models that can read},
  author={Singh, Amanpreet and Natarajan, Vivek and Shah, Meet and Jiang, Yu and Chen, Xinlei and Batra, Dhruv and Parikh, Devi and Rohrbach, Marcus},
  booktitle={Proc. of IEEE/CVF CVPR},
  pages={8317--8326},
  year={2019}
}

@inproceedings{sqa,
  title={Learn to explain: Multimodal reasoning via thought chains for science question answering},
  author={Lu, Pan and Mishra, Swaroop and Xia, Tanglin and Qiu, Liang and Chang, Kai-Wei and Zhu, Song-Chun and Tafjord, Oyvind and Clark, Peter and Kalyan, Ashwin},
  booktitle={Proc. of NeurIPS},
  volume={35},
  pages={2507--2521},
  year={2022}
}

@inproceedings{gqa,
  title={Gqa: A new dataset for real-world visual reasoning and compositional question answering},
  author={Hudson, Drew A and Manning, Christopher D},
  booktitle={Proc. of IEEE/CVF CVPR},
  pages={6700--6709},
  year={2019}
}

@misc{qwen25_vl,
    title = {Qwen2.5-VL},
    url = {https://qwenlm.github.io/blog/qwen2.5-vl/},
    author = {Qwen Team},
    year = {2025}
}

@inproceedings{pytorch,
  title={Pytorch: An imperative style, high-performance deep learning library},
  author={Paszke, Adam and Gross, Sam and Massa, Francisco and Lerer, Adam and Bradbury, James and Chanan, Gregory and Killeen, Trevor and Lin, Zeming and Gimelshein, Natalia and Antiga, Luca and others},
  booktitle={Proc. of NeurIPS},
  volume={32},
  pages={8024--8035},
  year={2019}
}

@inproceedings{vflowopt,
  title={Vflowopt: A token pruning framework for lmms with visual information flow-guided optimization},
  author={Yang, Sihan and Xu, Runsen and Cui, Chenhang and Wang, Tai and Lin, Dahua and Pang, Jiangmiao},
  booktitle={Proc. of IEEE/CVF ICCV},
  pages={23924--23934},
  year={2025}
}

@inproceedings{hidrop,
  title={Hidrop: Hierarchical vision token reduction in mllms via late injection, concave pyramid pruning, and early exit},
  author={Wu, Hao and Fan, Yingqi and Dai, Jinyang and Tong, Junlong and Ma, Yunpu and Shen, Xiaoyu},
  booktitle={Proc. of ICLR},
  year={2026}
}

@article{todre,
  title={Todre: Visual token pruning via diversity and task awareness for efficient large vision-language models},
  author={Li, Duo and Yang, Zuhao and Lu, Shijian},
  journal={arXiv preprint arXiv:2505.18757},
  pages={arXiv--2505},
  year={2025}
}

@inproceedings{tamp,
  title={Tamp: Token-adaptive layerwise pruning in multimodal large language models},
  author={Lee, Jaewoo and Xuan, Keyang and Ekbote, Chanakya and Polisetty, Sandeep and Fung, Yi R and Liang, Paul Pu},
  booktitle={Findings of ACL},
  pages={6892--6908},
  year={2025}
}

@article{swiftvlm,
  title={SwiftVLM: Efficient Vision-Language Model Inference via Cross-Layer Token Bypass},
  author={Qian, Chen and Yu, Xinran and Li, Danyang and Chi, Guoxuan and Yang, Zheng and Ma, Qiang and Miao, Xin},
  journal={arXiv preprint arXiv:2602.03134},
  year={2026}
}

@inproceedings{fitprune,
  title={Fit and prune: Fast and training-free visual token pruning for multi-modal large language models},
  author={Ye, Weihao and Wu, Qiong and Lin, Wenhao and Zhou, Yiyi},
  booktitle={Proc. of AAAI},
  volume={39},
  number={21},
  pages={22128--22136},
  year={2025}
}

@inproceedings{holitom,
  title={HoliTom: Holistic Token Merging for Fast Video Large Language Models},
  author={Shao, Kele and Tao, Keda and Qin, Can and You, Haoxuan and Sui, Yang and Wang, Huan},
  booktitle={Proc. of NeurIPS},
  volume={38},
  pages={135547--135570},
  year={2026}
}

@inproceedings{dycoke,
  title={Dycoke: Dynamic compression of tokens for fast video large language models},
  author={Tao, Keda and Qin, Can and You, Haoxuan and Sui, Yang and Wang, Huan},
  booktitle={Proc. of IEEE/CVF CVPR},
  pages={18992--19001},
  year={2025}
}

@inproceedings{edge1,
  title={Towards efficient edge learning for large models in heterogeneous resource-limited environments},
  author={Liu, Defang and Wang, Zhibo and Pang, Xiaoyi and Sun, Yunan and Hu, Jiahui and Sun, Peng and Hu, Yuke},
  booktitle={Proc. of IEEE BigCom},
  pages={223--230},
  year={2023},
}

@article{edge2,
  title={AFLoRA: Adaptive Federated Fine-Tuning of Large Language Models with Resource-Aware Low-Rank Adaption},
  author={Zhou, Yajie and Pang, Xiaoyi and Wang, Zhibo},
  journal={arXiv preprint arXiv:2505.24773},
  year={2025}
}

@inproceedings{edge3,
  title={Towards online privacy-preserving computation offloading in mobile edge computing},
  author={Pang, Xiaoyi and Wang, Zhibo and Li, Jingxin and Zhou, Ruiting and Ren, Ju and Li, Zhetao},
  booktitle={Proc. of IEEE INFOCOM},
  pages={1179--1188},
  year={2022},
}

@article{edge4,
  title={Cost-efficient and secure federated learning for edge computing},
  author={Zhang, Zhuangzhuang and Wu, Libing and Wang, Zhibo and Hu, Jiahui and Ma, Chao and Liu, Qin},
  journal={IEEE Transactions on Mobile Computing},
  volume={24},
  number={12},
  pages={13615--13632},
  year={2025},
}

\end{document}